\documentclass[10pt]{article} 
\usepackage[preprint]{tmlr}

\usepackage{amsmath,amsfonts,bm}

\def\eqref#1{equation~\ref{#1}}

\def\1{\bm{1}}

\DeclareMathAlphabet{\mathsfit}{\encodingdefault}{\sfdefault}{m}{sl}
\SetMathAlphabet{\mathsfit}{bold}{\encodingdefault}{\sfdefault}{bx}{n}

\usepackage{hyperref}
\usepackage{url}
\usepackage{booktabs}       
\usepackage{amsfonts}       
\usepackage{nicefrac}       
\usepackage{graphicx}
\usepackage{amsmath}
\usepackage{amssymb}
\usepackage{xcolor}         
\usepackage{lipsum}
\usepackage{tabularx}
\newcolumntype{Y}{>{\centering\arraybackslash}X}
\newcolumntype{P}{>{\centering\arraybackslash}p}
\newcolumntype{L}{>{\raggedright\arraybackslash}X}
\usepackage{multirow}
\usepackage[flushleft]{threeparttable}
\usepackage{subcaption}
\usepackage{natbib}
\usepackage{algorithm}      
\usepackage[noend]{algpseudocode} 
\floatname{algorithm}{Algorithm}

\usepackage{caption}
\usepackage{tikz}

\newcommand{\TMIX}[0]{\textit{TransformMix }}
\newcommand{\TMIXb}[0]{\textit{TransformMix}}
\newcommand*\emptycirc[1][1ex]{\tikz\draw (0,0) circle (#1);} 
\newcommand*\halfcirc[1][1ex]{%
  \begin{tikzpicture}
  \draw[fill] (0,0)-- (90:#1) arc (90:270:#1) -- cycle ;
  \draw (0,0) circle (#1);
  \end{tikzpicture}}
\newcommand*\fullcirc[1][1ex]{\tikz\fill (0,0) circle (#1);} 
\newcommand*\f[1]{\ifcase#1 \emptycirc\or\halfcirc\or\fullcirc\fi}

\title{TransformMix: Learning Transformation and Mixing Strategies from Data}

\author{\name Tsz-Him Cheung \email thcheungae@connect.ust.hk \\
      \addr Department of Computer Science and Engineering\\
      Hong Kong University of Science and Technology
      \AND
      \name Dit-Yan Yeung \email dyyeung@cse.ust.hk \\
      Department of Computer Science and Engineering\\
      \addr Hong Kong University of Science and Technology
}

\def\month{MM}  
\def\year{YYYY} 
\def\openreview{\url{https://openreview.net/forum?id=XXXX}} 

\begin{document}

\maketitle

\begin{abstract}
Data augmentation improves the generalization power of deep learning models by synthesizing more training samples. Sample-mixing is a popular data augmentation approach that creates additional data by combining existing samples. Recent sample-mixing methods, like Mixup and Cutmix, adopt simple mixing operations to blend multiple inputs. Although such a heuristic approach shows certain performance gains in some computer vision tasks, it mixes the images blindly and does not adapt to different datasets automatically. A mixing strategy that is effective for a particular dataset does not often generalize well to other datasets. If not properly configured, the methods may create misleading mixed images, which jeopardize the effectiveness of sample-mixing augmentations. In this work, we propose an automated approach, \TMIXb, to learn better transformation and mixing augmentation strategies from data. In particular, \TMIX applies learned transformations and mixing masks to create compelling mixed images that contain correct and important information for the target tasks. We demonstrate the effectiveness of \TMIX on multiple datasets in transfer learning, classification, object detection, and knowledge distillation settings. Experimental results show that our method achieves better performance as well as efficiency when compared with strong sample-mixing baselines.
\end{abstract}

\section{Introduction} \label{sec:introduction}
\noindent Modern deep learning models achieve remarkable success in important computer vision tasks, like object classification~\citep{alexnet,resnet,wrn} and semantic segmentation~\citep{fasterrcnn,yolov4}. Despite these reported successes, deep learning models can easily overfit when the training set is quantitatively deficient. To generalize deep learning models beyond finite training sets, data augmentation is a widely adopted approach that synthesizes additional samples to expand the training sets~\citep{da}. Conventional data augmentation applies pre-defined image processing functions, such as random cropping, flipping and color adjustment, to create additional views of the same data. To reduce the manual effort in searching for the appropriate augmentation configuration, automated data augmentation (AutoDA) methods have been proposed to search for the optimal augmentation policy for a dataset~\citep{aa, pba, fastaa, modals, adaaug}. Given the right choice of augmentation, these transformation-based techniques induce constructive inductive biases in the dataset, thereby improving the generalization power of machine learning models. 

Sample-mixing is a different data augmentation approach that synthesizes additional samples by combining multiple images. Unlike conventional data augmentation, sample-mixing does not require the specification of domain-specific transformations, allowing it to be flexibly deployed to other data domains. The seminal work Mixup interpolates two training images with their one-hot-encoded label proportionally~\citep{mixup}, while CutMix randomly replaces a patch of an image with another image~\citep{cutmix}. Although these mixing strategies can bring slight improvements in some computer vision tasks~\citep{yolov4}, images are combined without considering their content. Consequently, the mixed images may contain misleading training signals and undermine the model performance. Specifically, Mixup may lead to the manifold collision problem where an interpolated image sits on the real data manifold~\citep{adamixup}; the random replacement in CutMix may remove important regions that are crucial in identifying an object instance in classification tasks. To this end, recent works apply additional saliency constraints to avoid the crucial information being removed during the mixing process~\citep{puzzlemix, supermix, comixup, automix, smix}.

Inspired by recent AutoDA advancements and saliency-aware mixing methods, we develop an automated method, named \TMIXb, that learns a better mixing strategy from a dataset with two criteria: (1) the mixing strategy should create mixed outputs that maximally preserve the visual saliency of the input images; (2) the mixing strategy should be learned from the dataset automatically. For more practical usage in transfer learning, we also investigate whether a discovered mixing strategy can be transferred to create new augmented images on unseen datasets. At a high level, \TMIX exploits self-supervision signals from a pre-trained teacher network to learn a mixing module for predicting the transformations and mixing masks to create mixed images that preserve the salient information of the input images. To summarize, we make four major contributions.

\begin{itemize}
	\item We propose \TMIXb, which employs a novel mixing module that predicts the transformations and mixing masks to create more advantageous mixed images with maximal preservation of visual saliency.
	\item We introduce an automated framework that trains the mixing module on a given dataset efficiently.
	\item We demonstrate that our method improves state-of-the-art results on several popular classification benchmarks and achieves $4\times$ to $18\times$ speed-up compared to other saliency-aware sample-mixing baselines.
	\item We demonstrate that our method can be transferred to augment new unseen datasets. The transferred method provides non-trivial improvements over other sample-mixing methods. We also show the effectiveness of \TMIX on classification, object detection, and knowledge distillation tasks.
\end{itemize}

\begin{figure}[t]
\centering
\includegraphics[width=0.7\linewidth]{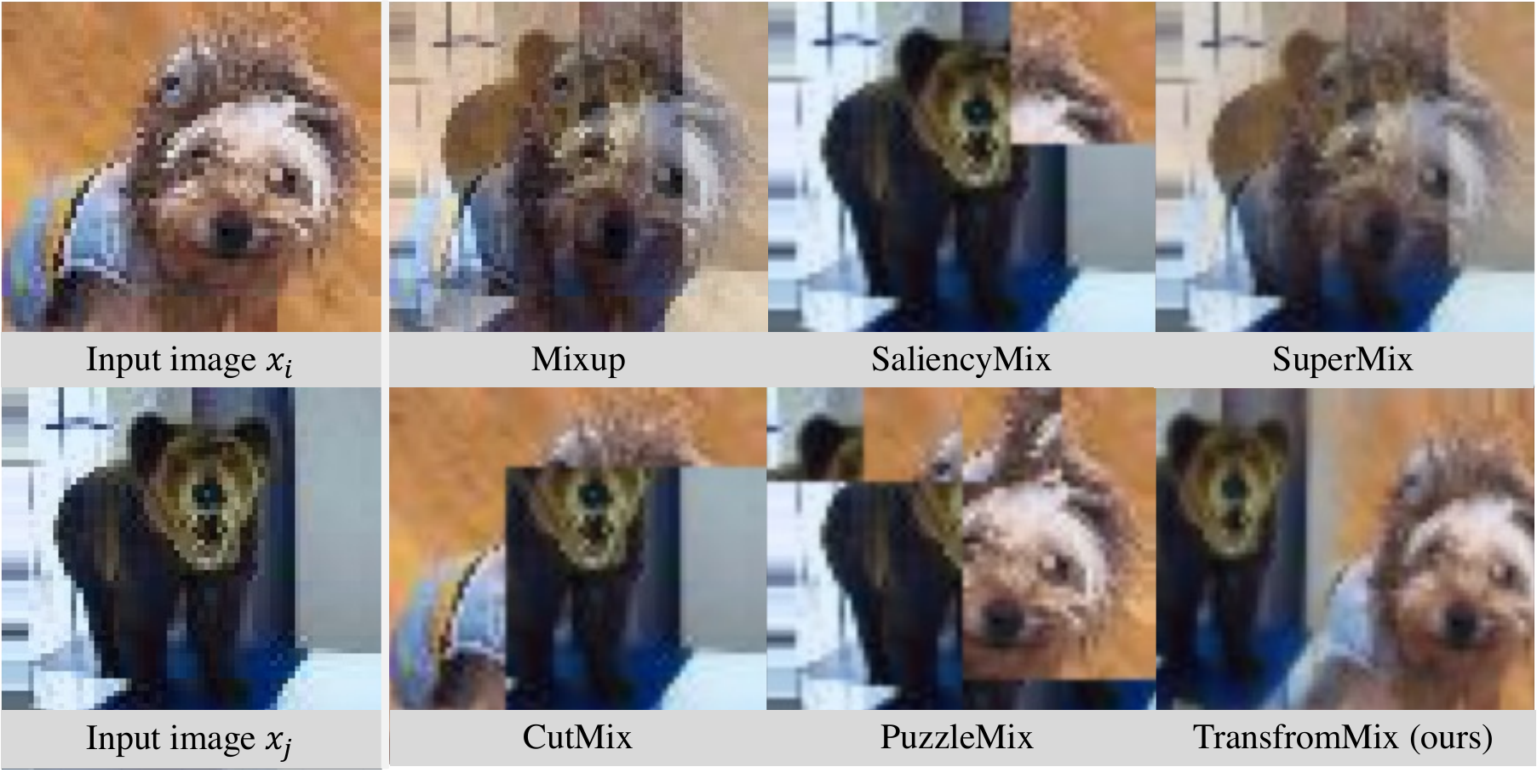}
\caption{Visual comparison of different sample-mixing methods on a dog and bear image. Our method better preserves the important region of the input images.}
\label{fig:demo}
\end{figure}

\section{Related Works} \label{sec:related_works}
\noindent \textbf{Automated Data Augmentation.} Conventional data augmentation applies label-preserving transformations to augment data. However, the specification and choice of data augmentation functions rely heavily on expert knowledge or repeated trials. To address this shortcoming, AutoDA learns the optimal policy to augment the target dataset~\citep{autoda}. AutoAugment takes a reinforcement learning approach to learn the probability and magnitude of applying different transformations to the target dataset~\citep{aa}. PBA~\citep{pba} and RandAugment~\citep{randaug} study more efficient search methods to reduce the expensive search effort of AutoAugment. Inspired by AutoAugment, AdaAug~\citep{adaaug} learns an adaptive augmentation strategy for each data instance, while MODALS~\citep{modals} learns the optimal augmentation strategy to apply four transformations in the latent space, thereby allowing the method to be deployed to multiple modalities. Despite the successes reported by these AutoDA methods, they require domain knowledge in designing domain-specific label-preserving transformations. Moreover, they have not studied the optimal way to mix multiple images.

\textbf{MixUp.} MixUp~\citep{mixup} uses the convex combination of input images as new augmented images: $x' = \lambda x_1 + (1-\lambda) x_2$, where $x_1$ and $x_2$ are two training images, $\lambda$ is the mixing coefficient and $x'$ is the augmented image. Following Mixup, Manifold Mixup~\citep{manifoldmixup} applies Mixup to mix the latent representations, while CutMix~\citep{cutmix} replaces a random patch of an image with another image. These techniques disregard the image content and may dilute or occlude the salient information, which is crucial for the target task. 

\textbf{Saliency-aware Sample-mixing.} To mitigate the mentioned problem, several recent works attempted to create more advantageous mixed images by preserving the salient information in the mixed image. SaliencyMix~\citep{smix} and ResizeMix~\citep{resizemix} detect the saliency information and prevents CutMix from replacing the image patch that contains rich information. SuperMix utilizes a teacher model to optimize some mixing masks applied to the input images: $x' = m_1 \odot x_1 + m_2 \odot x_2$, where $m_1$ and $m_2$ denote the mixing masks and $\odot$ denotes the elementwise multiplication. In AutoMix~\citep{automix}, the mixing masks are computed by a mix block based on the image features. However, consider two input images having the salient information at the same pixel location, for example, the face of a dog in input image $x_i$ and the face of a bear in input image $x_j$ as illustrated in Figure~\ref{fig:demo}, optimizing a mixing mask alone cannot prevent the two faces being blended in the mixed results. Hence, PuzzleMix~\citep{puzzlemix} optimizes the mixing masks together with transport plans $\Pi_1$ and $\Pi_2$ that specify the mass to be transported between different pixel locations: $x' = m_1 \odot \Pi_1 x_1 + m_2 \odot \Pi_2 x_2$. Although the method avoids the overlapping of the salient regions to a certain degree, the method splits the input images into different blocks and shifts their locations. This creates puzzle-like artifacts in the resulting image, violating the natural image prior~(see Figure~\ref{fig:demo}). We summarize the features of various sample-mixing methods and highlight the strength of \TMIX in Table~\ref{tab:compare}. 

\begin{table*}[ht]
  \caption{Comparison between the properties of various sample-mixing data augmentation methods.}
  \label{tab:compare}
  \small
  \centering
  \begin{tabularx}{\textwidth}{LY*{6}{c}}
    \toprule
        & Mixup & Cutmix & SaliencyMix & PuzzleMix & SuperMix & \TMIXb \\
    \midrule
        Saliency-aware mixing           & \f{0} & \f{0} & \f{2} & \f{2} & \f{2} & \f{2} \\ \midrule
        Saliency-aware transformation   & \f{0} & \f{0} & \f{1} & \f{1} & \f{0} & \f{2} \\ \midrule
        Natural blend of images         & \f{0} & \f{0} & \f{0} & \f{0} & \f{2} & \f{2} \\ \midrule
        Effective for image 
        classification                  & \f{1} & \f{1} & \f{1} & \f{2} & \f{2} & \f{2} \\ \midrule
        Fast inference time             & \f{2} & \f{2} & \f{1} & \f{0} & \f{0} & \f{1} \\ \midrule
        Transferable augmentation policy         & \f{0} & \f{0} & \f{0} & \f{0} & \f{0} & \f{2} \\
    \bottomrule
  \end{tabularx}
  \begin{tablenotes}
    \item \hfil$\f2=\text{provides property}$; $\f1=\text{partially provides property}$; $\f0=\text{Not tested or does not provide property}$
  \end{tablenotes}
\end{table*}

\TMIX aims to learn the transformation and mixing strategies that generate more advantageous mixed data automatically. Learning such strategies poses two major challenges. First, a mixing strategy needs to decide the output of every pixel location, which is harder than learning an augmentation policy from a set of 10 to 20 augmentation functions in previous AutoDA works. Therefore, formulating the transformation and mixing strategy and optimizing it efficiently is a challenging problem. Second, previous works create mixed input with unnatural mixing boundaries or require manual specification of additional constraints to create more realistic outputs. Designing an automated approach to generate more natural blends of images is non-trivial. Hence, developing a transferable mixing strategy like \TMIX to reduce the computation efforts on new datasets is a favorable solution for data practitioners.

\section{Methodology} \label{sec:method}
\begin{figure}[h]
  \centering
  \includegraphics[width=0.6\linewidth]{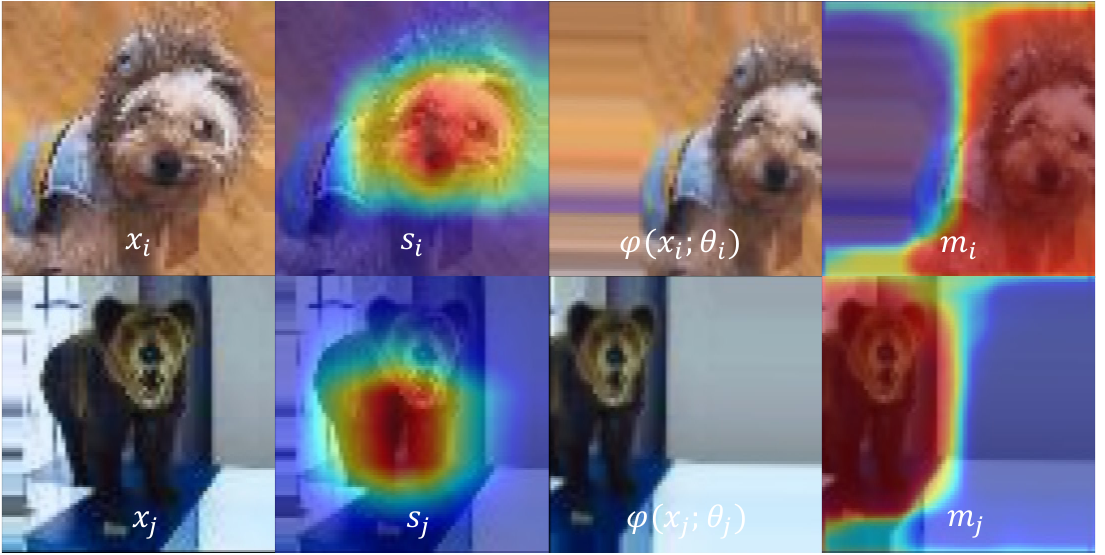}
  \caption{Illustrations of the intermediate results during mixing. From left to right, the columns show the visualizations of the input images, CAMs, transformed images and predicted masks.}
  \label{fig:test1}
\end{figure}

\noindent \TMIX first learns a mixing strategy from a dataset under the supervision of a pre-trained teacher model and then creates mixed data to train some end task networks. We formulate the mixing strategy by predicting some transformations and mixing masks applied to the input images. Given two distinct instances $(x_i, y_i)$ and $(x_j, y_j)$, where $x_i\in \mathbb{R}^{C\times W \times H}$ is the $i$-th training sample with class label $y_i$, $C$ channels and $W \times H$ dimension, \TMIX aims to find the 
effective transformations $\phi_i, \phi_j$ and mixing masks $m_i, m_j \in [0,1]^{W \times H}$ applied to $x_i$ and $x_j$ for creating better mixed images $x'$ according to:
\begin{equation} \label{eq:general}
	x' = m_i \odot \phi_i(x_i) + m_j \odot \phi_j(x_j), \,\,\,
	y' = \lambda y_i + (1-\lambda)y_j,
\end{equation}
where $\lambda$ is a mixing coefficient for mixing $x_i$ and $x_j$. Under this formulation, the transformations help to separate the salient regions even if they completely overlap in the input images, while the mixing masks help to reveal the more important regions of the candidate images with respect to the target task. To better preserve visual saliency, \TMIX utilizes the class activation maps (CAMs) of the input images when predicting the mixing strategy. The intermediate visual illustrations of the input images, saliency maps, transformations and predicted mixing masks are presented in Figure~\ref{fig:test1}. In the following section, we explain the details of each component and then the training framework of \TMIXb.

\begin{figure}[t]
\centering
\includegraphics[width=\textwidth]{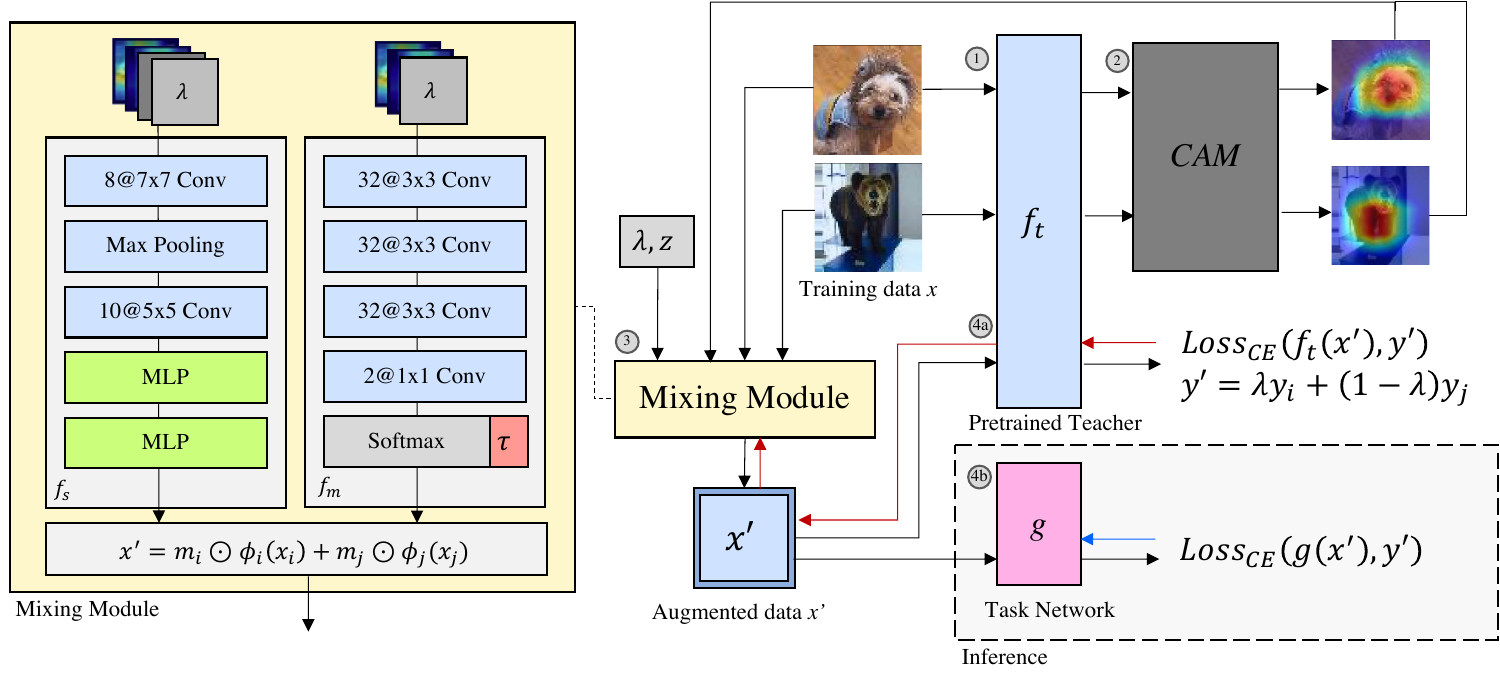}
\caption{Overview of \TMIXb. The black arrows indicate the forward pass; the red arrows indicate the gradient flow when training the spatial transformation network $f_s$ and mask prediction network $f_m$; the blue arrow indicates the gradient flow when training the task network $g$. \textbf{Step~1\&2}: the CAMs of the input images $(x_i, x_j)$ are extracted from the pre-trained teacher $f_t$. \textbf{Step 3}: the CAMs, input images and sampled mixing coefficient $\lambda$ are supplied to the mixing module to compute the transformations $(\phi_i, \phi_j)$ and mixing masks $(m_i, m_j)$  to create the mixed image $x'$, which is used in \textbf{step 4a}: training the mixing module, or \textbf{step 4b}: training the the task network.}
\label{fig:method}
\end{figure}

\subsection{TransformMix}
\noindent \TMIX comprises three components: a pre-trained teacher model, a saliency detector and a mixing module. \TMIX first trains the mixing module under the supervision of the teacher model. Once the mixing module is trained, it is used to create mixed images to train a new task network. The overview of \TMIX is shown in Figure~\ref{fig:method}.

\textbf{Saliency Detector.}
It was previously suggested that considering the image saliency during mixing can improve the quality of mixed images~\citep{puzzlemix,smix}. In our method, we adopt a neural network approach, which takes visual saliency as input features to make a prediction of a mixing strategy. We propose to condition on the visual saliency rather than the image or latent image representation because visual saliency is a generic measure applicable to all images, whereas the image or latent image representation is mostly restricted to a dataset. Therefore, learning from visual saliency facilitates transferring the trained mixing module to unseen data. 

Typically, a saliency detector generates a heatmap $s_i \in [0,1]^{W \times H}$ that highlights the important regions of $x_i$. In practice, we can use any saliency detection algorithms or explainable AI methods to extract salient information from input images. As \TMIX employs a pre-trained teacher network in the later training stage, we exploit the readily available pre-trained weights and use CAMs~\citep{cam} to estimate the salient regions of an image. In essence, CAMs are the summation of global averaged features weighted by the weights learned by the classification layer. The pixel location that is more important to the classification task will be assigned a larger value in the heatmap.

\textbf{Spatial Transformer Network.}
Unlike PuzzleMix~\citep{puzzlemix} which divides the input images into multiple blocks and re-organizes them to avoid overlapping salient regions, \TMIX encompasses a more comprehensive set of transformations and creates mixed outputs complying with the natural image prior. As a learnable transformation method, Spatial Transformer Network~(STN) is a convolutional neural network module that spatially transforms an image by predicting the appropriate transformation for each data instance~\citep{stn}. Different from the previous works that predict the transformation based on a single image, we propose a novel approach using an STN to predict 6 affine parameters for each of the two input images based on their CAMs, a mixing coefficient $\lambda$ and a sampled noise $z$. The mixing coefficient $\lambda$ allows more emphasis on different proportions of mixing and is sampled from the Beta distribution with parameter $\alpha$. Sampled from the Normal distribution, $z$ adds certain diversity when transforming the images. The mixing coefficient and sampled noise are resized and appended to the CAMs as additional channels. We use $f_s: \mathbb{R}^{4 \times W \times H} \rightarrow \mathbb{R}^{2 \times 6}$ to denote the STN and $\theta$ to denote the predicted affine parameter, which is given by:
\begin{equation}
\begin{aligned}
	\theta_i, \theta_j = f_s(s_i, s_j, \lambda, z), \lambda \sim \text{Beta}(\alpha, \alpha), z \sim N(0,I)
\end{aligned}
\end{equation}
Using the affine parameters predicted by $f_s$, the transformations are then performed on the input images accordingly to avoid overlapping of salient regions.  Although recent AutoDA methods can search in a larger set of transformations, like color adjustments and other non-differentiable operations, the spatial transformations, including scaling, cropping, rotations, as well as non-rigid deformations, are sufficient to tackle the saliency overlapping issue. Therefore, we opt for the differentiable STN approach, which can be optimized efficiently using gradient-based methods, instead of costly AutoDA search methods, to characterize $\phi$ in \eqref{eq:general}.

\begin{figure}[h]
  \centering
  \includegraphics[width=0.6\linewidth]{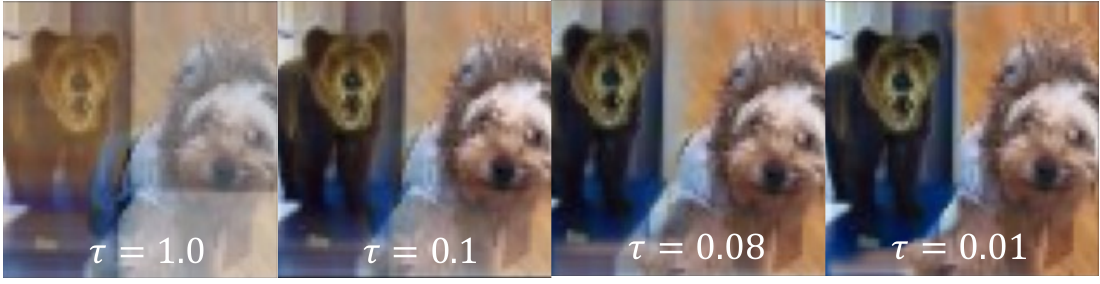}
  \caption{Illustrations of the effect when using different temperature values $\tau$. The third image is the mixed result with learned $\tau=0.08$.}
  \label{fig:test2}
\end{figure}

\textbf{Mask Prediction Network.}
The mask prediction network $f_m$ receives the transformed CAMs and mixing coefficient $\lambda$ to predict the mixing masks $(m_i, m_j)$. We implement $f_m$ as a spatial-preserving convolutional neural network, i.e., $f_m: \mathbb{R}^{3 \times W \times H} \rightarrow [0,1]^{2 \times W \times H}$.  The softmax function is applied to the output layer to ensure that the mixing masks sum to 1 at each pixel location $(w,h)$. In addition, a learnable temperature parameter $\tau$ is introduced to control the smoothness of the mixing boundary. Specifically, a lower temperature value will result in a sharper blending boundary~(see Figure~\ref{fig:test2}). With $o$ denoting the hidden feature before the softmax layer and $\varphi(\cdot; \theta)$ denoting the affine transformation with parameter $\theta$, the mixing masks are computed as:
\begin{equation}
	m_i, m_j = f_m(\varphi(s_i;\theta_i), \varphi(s_j;\theta_j), \lambda),\;\;\;\;\;
	m_i^{(w,h)} = 
	\frac	{\exp (o_i^{(w,h)}/\tau)}
			{\sum_{k \in \{i,j\}}\exp (o_k^{(w,h)}/\tau)}
\end{equation}

\subsection{Training}
\noindent \TMIX employs a two-stage training process. At the search stage, the mixing module is trained to learn the transformation and mixing strategy from a dataset. At inference time, the learned mixing module will generate new mixed images for training a new classifier on the dataset.

In AutoDA, augmentation policies are optimized with the goal to maximize the validation performance on a hold-out dataset directly. The process requires the training of multiple child models, which is computationally expensive to achieve. Specifically, AutoAugment spends over 5,000 GPU hours learning the augmentation policy on CIFAR-10~\citep{aa}. To this end, we adopt a surrogate objective that correlates to the effectiveness of mixed training images. Inspired by SuperMix~\citep{supermix}, we utilize a pre-trained teacher model $f_t$ to guide the learning of the mixing module. Specifically, we update the mixing module to minimize the classification loss on the mixed input $x'$ with label $y'$ in \eqref{eq:mixing_process}. With the supervision signals given by the pre-trained network, the mixing module ($f_s, f_m$) learns to construct mixed images such that $f_t$ can uncover the constituting objects in $x'$. This encourages the mixing module to fit more important information in the mixed results, avoiding the salient information being diluted implicitly. 
\begin{equation} \label{eq:mixing_process}
	x' = m_i \odot \varphi(x_i;\theta_i) + m_j \odot \varphi(x_j;\theta_j), \,\,\,
	y' = \lambda y_i + (1-\lambda)y_j
\end{equation}

At inference time, the trained mixing module creates mixed images to train a new classifier following the standard model training process. In the previous sections, we explained the details of \TMIX to mix two input images. In practice, \TMIX can be generalized to mix more images by replacing the Beta distribution with Dirichlet distribution and modifying the input and output sizes of $f_s$ and $f_m$ accordingly. The general training step to mix $k$ images by \TMIX is depicted in Algorithm~\ref{alg:tmix}.

\label{app:algor}
\begin{algorithm}
\begin{algorithmic}[1]
\caption{Training of \TMIX}
\label{alg:tmix}
\Procedure{Train}{input images $\{x_i\}_{i=1}^k$, teacher network $f_{t}$, spatial transformer network ${f_s}$, mask prediction network $f_{m}$, task network $g$, mixing parameter $\alpha$}
\State $S = \{\mathit{CAM}(x_i;f_t)\}_{i=1}^k$
\State Sample $\Lambda=(\lambda_1, \ldots, \lambda_{k})$ from Dir($\alpha$)
\State Sample $z$ from N($0, I$)
\State $\Theta = f_s(S, \Lambda, z)$
\State $M = f_m( \{ \varphi(s_i;\theta_i)\}_{i=1}^k, \Lambda)$
\State $x' = \sum_{i=1}^{k} m_i \odot \varphi(x_i;\theta_i)$
\State $y' = \sum_{i=1}^{k} \lambda_i y_i$
\If{ $\mathit{SEARCH}$ } \textit{\textcolor{black!60}{\# Training the mixing module}}
	\State $\mathcal{L} = \mathit{CrossEntropyLoss}(f_t(x'), y')$
	\State Update $f_s, f_m$
\Else \,\textit{\textcolor{black!60}{\# Training the task network}}
	\State $\mathcal{L} = \mathit{CrossEntropyLoss}(g(x'), y')$
	\State Update $g$
\EndIf
\EndProcedure
\end{algorithmic}
\end{algorithm}

\section{Experiments and Results}
\subsection{Experiment Setup}
\noindent \textbf{Mixing Module}. The mixing module comprises a spatial transformer network and a mask prediction network. We follow the implementation of the spatial transformer network~\citep{stn} except the last layer is modified for additional outputs to predict the transformation parameters for two images. The mask prediction network has three convolution layers with 3$\times$3 kernel size, 32 channels and ReLU activation, followed by a 1$\times$1 convolution layer. Padding is applied to ensure that the mixing masks share the same spatial dimension as the input images. The mixing module is trained using SGD with a 0.0005 learning rate and 0.01 weight decay for 100 epochs using a batch size of 128.

\noindent \textbf{Teacher Network}. In our experiments, the teacher networks are trained on the target datasets following the simple baseline methods. Specifically, the teacher network uses the same network architecture and is trained using the same configurations as the task networks. The test accuracy of the teacher networks on each dataset is the same as the ``Simple'' baseline listed in Table~\ref{tab:transfer2} and~\ref{tab:direct}. 

\subsection{Transfer Classification}
\noindent We study whether a mixing module trained on a dataset can be exploited to create effective mixed data for other datasets. We consider a realistic scenario where a pre-trained classifier is fine-tuned on some downstream tasks. We follow the transfer setting in~\citep{adaaug} to fine-tune an ImageNet pre-trained network on four classification datasets: Oxford Flowers~\citep{flower}, Oxford-IIIT Pets~\citep{pet}, FGVC Aircraft~\citep{aircraft} and Stanford Cars~\citep{car}.

Fine-tuning ImageNet pre-trained model is one of the most promising transfer-learning techniques nowadays. Here, we use ResNet-50 with pre-trained IMAGENET1K\_V1 weights provided by \texttt{torchvision}\footnote{\url{https://pytorch.org/vision/main/models/generated/torchvision.models.resnet50.html}}. We extract the CAMs of input images from the pre-trained model and fine-tune the model on four downstream datasets: Oxford Flowers~\citep{flower}, Oxford-IIIT Pets~\citep{pet}, FGVC Aircraft~\citep{aircraft} and Stanford Cars~\citep{car}. Compared to Tiny-ImageNet and ImageNet, these datasets have a larger number of classes but few samples per class.

We apply the mixing module pre-trained on TinyImageNet for the TransformMix baseline. For the other baselines, we use the optimal hyperparameters for Tiny-ImageNet suggested by the original papers. The pre-trained model is fine-tuned on the downstream datsets for 100 epochs using a learning rate of 0.001 and batch size of 64. The results in Table~\ref{tab:transfer2} demonstrate that \TMIX outperforms Mixup, CutMix, SaliencyMix and PuzzleMix on the Pet, Car and Aircraft datasets and achives similar performance on the Flower datsaet in terms of top-1 and top-5 accuracy.
\begin{table}[h]
  \caption{Test-set Top-1 / Top-5 accuracy (\%) for fine-tuning a pre-trained network to the Flower, Pet, Car and Aircraft datasets using the mixing module learned from CIFAR-100.}
  \label{tab:transfer2}
  \centering
  \begin{tabularx}{\textwidth}{lYYYY}
    \toprule
     Method & Flower & Pet & Car & Aircraft \\
    \midrule
	 Simple 
	 & 97.55 / \textbf{99.87} & 92.88 / 99.48 & 79.43 / 95.52 & 68.79 / 92.34 \\
	 Mixup
	 & 97.92 / 99.75 & 93.04 / 99.31 & 80.84 / 95.77 & 70.83 / 93.00 \\
     CutMix      
     & 97.67 / 99.63 & 92.09 / 99.20 & 80.91 / 95.94 & 69.63 / 93.06 \\
     SaliencyMix  
     & 98.16 / 99.75 & 92.83 / 99.34 & 78.90 / 95.32 & 69.42 / 92.25 \\
     PuzzleMix   
     & \textbf{98.28} / 99.75 & 92.66 / \textbf{99.50} & 80.62 / 95.87 & 70.89 / 92.97 \\
     \TMIXb 
     & 98.16 / \textbf{99.87} & \textbf{93.34} / \textbf{99.50} & \textbf{82.11} / \textbf{96.48} & \textbf{72.36} / \textbf{93.12} \\
     (ours) 
     & \small{$\pm$0.06} / \small{$\pm$0.06} & \small{$\pm$0.13} / \small{$\pm$0.04} & \small{$\pm$0.47} / \small{$\pm$0.18} & \small{$\pm$0.35} / \small{$\pm$0.16} \\
     \bottomrule
  \end{tabularx}
\end{table}

\begin{table}[h]
  \caption{Test-set Top-1 / Top-5 accuracy (\%) on CIFAR-100, Tiny-ImageNet and ImageNet datasets.}
  \label{tab:direct}
  \centering
  \begin{tabularx}{\textwidth}{lYYYY}
    \toprule
       &
 	  \multicolumn{2}{c}{CIFAR-100} &
      Tiny-ImageNet &
      ImageNet \\
      \cmidrule{2-3} \cmidrule{4-4} \cmidrule{5-5}
      Method & 
      WRN-$28\times 10$ & PreActResNet-18 &
      PreActResNet-18 & 
      ResNet-50 \\
    \midrule
    Simple             & 78.86 / 93.67   & 76.33	 / 91.02   & 57.23 / 73.65   & 75.69 / 92.66   \\
    Mixup        & 81.73 / 95.02   & 76.84 / 92.42   & 56.59 / 73.02   & 77.01	 / 93.52   \\
    Manifold MixUp     & 82.60 / 95.63   & 79.02 / 93.37   & 58.01 / 74.12   & 76.85 / 93.50   \\
    CutMix             & 82.50 / 95.31   & 76.80 / 91.91   & 56.67 / 75.52   & 77.08 / 93.45   \\
    AugMix             & 79.56 / 94.26   & 75.31 / 91.62   & 55.97 / 74.68   & 76.75 / 93.30   \\
    PuzzleMix         & 84.05 / 96.08   & 80.38	 / 94.15   & 63.48 / 81.05   & 77.51 / 93.76   \\
    SuperMix          & 83.60 / \,\;\;\;-\;\;\;\,       & -	 / -           & - / -           & \textbf{77.60} / 93.70   \\
    \midrule
    \TMIXb & \textbf{84.07} / \textbf{96.97}   & \textbf{80.39} / \textbf{95.37}   & \textbf{65.72} / \textbf{85.01}   & \textbf{77.60}	 / \textbf{93.89}   \\ 
    (ours) & \small{$\pm$0.05} / \small{$\pm$0.07} & \small{$\pm$0.04} / \small{$\pm$0.05} & \small{$\pm$0.16} / \small{$\pm$0.25} & \small{$\pm$0.04} / \small{$\pm$0.06}\\
    \bottomrule
  \end{tabularx}
\end{table}

\subsection{Direct Classification}
\noindent Following the line of sample-mixing research, we evaluate \TMIX on CIFAR-100~\citep{cifar10}, Tiny-ImageNet~\citep{tinyimagenet} and ImageNet~\citep{imagenet} using WideResNet-28$\times$10~\citep{wrn}, PreActResNet-18~\citep{presnet} and ResNet-50~\citep{resnet}. We compare our results with simple sample-mixing methods: Mixup~\citep{mixup}, Manifold Mixup~\citep{manifoldmixup}, CutMix~\citep{cutmix} and AugMix~\citep{augmix} as well as saliency-aware mixing methods: PuzzleMix~\citep{puzzlemix} and SuperMix~\citep{supermix}. In the direct classification experiment, we first train the mixing module (i.e., the spatial transformer network and mask prediction network) on the target dataset, and then utilize the mixing module to create new mixed images for training a task classifier from scratch.

\textbf{CIFAR-100}.
Following the experiment setting in PuzzleMix~\citep{puzzlemix}, we train WRN-28$\times$10~\citep{wrn} and PreActResNet-18~\citep{presnet} on CIFAR-100. We follow the training protocol in~\citep{puzzlemix} to train WRN-28$\times$10 for 400 epochs and PreActResNet-18 for 1200 epochs. We use an initial learning rate of 0.1 and decay it by a factor of 0.1 at the 200$^{th}$ and 300$^{th}$ epochs for WRN-28$\times$10 and at the 400$^{th}$ and 800$^{th}$ epochs for PreActResNet-18. We adopt the reported baseline performances from~\citep{puzzlemix}, which are tested under the same experiment setup. We repeat the experiments with three random seeds and report the averaged top-1 and top-5 test-set accuracy together with the standard deviation in Table~\ref{tab:direct}.

Experimental results show that our method achieves comparable top-1 accuracy as state-of-the-art methods. In addition, \TMIX achieves greater gains in the top-5 accuracy with 0.89\% and 1.22\% improvements over the best performed baseline on WRN-28$\times$10~\citep{wrn} and PreActResNet-18~\citep{presnet}, respectively. This provides evidence that \TMIX can preserve more object information, facilitating the task network to learn more effectively.

\textbf{Tiny-ImageNet}. Tiny-ImageNet contains 500 training images of 200 classes with a resolution of $64\times64$~\citep{tinyimagenet}. We follow PuzzleMix~\citep{puzzlemix} to train the PreActResNet18 network on the Tiny-ImageNet dataset for 1,200 epochs with an initial learning rate of 0.2 and decay it by a factor of 0.1 at the 600$^{th}$ and 900$^{th}$ epochs. As shown in Table~\ref{tab:direct}, \TMIX outperforms PuzzleMix by 2.24\% and 3.96\% in terms of top-1 and top-5 accuracy, respectively.

\textbf{ImageNet}. We also evaluate our methods on the ImageNet dataset with ResNet-50. The dataset contains 1,281,167 training images in 1,000 categories. Following the experiment setup prescribed in~\citep{puzzlemix}, we train ResNet-50 on the ImageNet for 100 epochs using an initial learning rate of 0.5 with learning rate warm-up and weight decay on resized ImageNet images. Similar to the CIFAR-100 experiment, our method achieves comparable top-1 accuracy and better top-5 accuracy than other baselines (see Table~\ref{tab:direct}).

\subsection{Object Detection}
\noindent This section compares \TMIX with other mixing baselines on the Pascal VOC object detection task~\citep{voc}. We follow SaliencyMix~\citep{smix} to use the ResNet-50 pre-trained on ImageNet with different mixing strategies as the backbone network of Faster RCNN~\citep{fasterrcnn}. We fine-tune the last 5 layers of the backbone networks with the Region Proposal Network and RoI Heads on Pascal VOC 2007 and 2012 data and test on the VOC 2007 test data. Following the same protocol, we fine-tune the model with a batch size of 8 and a learning rate of 0.0004 for 41,000 iterations. We decay the learning rate by a factor of 0.1 at the 33,000$^{th}$ iteration. As foreground objects are more important than the backgrounds in object detection tasks, saliency-preserving mixing strategies can create more advantageous augmented images to improve detection performance. Specifically, Table~\ref{tab:objectdetection} shows that \TMIX outperforms the simple baseline and leads other mixing methods in terms of the mAP score on the Pascal VOC object detection task.

\begin{table}[ht]
  \caption{Comparison of MixUp, SaliencyMix, PuzzleMix and \TMIX on Pascal VOC object detection task.}
  \label{tab:objectdetection}
  \centering
   \begin{tabularx}{\linewidth}{lYYYYY}
    \toprule
      & Simple & Mixup & SaliencyMix & PuzzleMix & \TMIXb \\
    \midrule
     Pascal VOC (mAP) & 74.2 & 75.2 & 75.3 & 75.6 & \textbf{75.7} \\
    \bottomrule
  \end{tabularx}
\end{table}

\subsection{Execution Time}
\noindent We compare the execution time of our method with the saliency-aware mixing methods: PuzzleMix and SuperMix. At inference time, our method uses a single forward pass to predict the mixing mask and transformation, while PuzzleMix and SuperMix iteratively compute the optimal mixing masks and transport plans. We test the average processing time to generate a batch of 128 mixed images for 10 trials. All the experiments are conducted 
using an NVIDIA RTX3090 GPU card. It is found that \TMIX is 3.61$\times$ faster than PuzzleMix and 3.97$\times$ faster than SuperMix on CIFAR-10 and 2.08$\times$ faster than PuzzleMix and 18.51$\times$ faster than SuperMix on ImageNet (see Table~\ref{tab:time}).

\begin{table}[h]
  \caption{Comparsion between the execution time of PuzzleMix, SuperMix and \TMIXb. The execution time of PuzzleMix and SuperMix is represented as the multiply of \TMIX execution time.}
  \label{tab:time}
  \centering
  \begin{tabularx}{\linewidth}{XYYY}
    \toprule
       &
      \TMIXb  &
      PuzzleMix & 
      SuperMix  \\
    \midrule
    CIFAR-10  & 1$\times$ & 3.61$\times$   & 3.97$\times$    \\
    ImageNet  & 1$\times$ & 2.08$\times$   & 18.51$\times$    \\
    \bottomrule
  \end{tabularx}
\end{table}

\subsection{Ablation Study}
\noindent In this section, we investigate the effectiveness of different components in \TMIXb. Specifically, we test \TMIX with only the STN and only the mask prediction network (MPN) on CIFAR-10 and CIFAR-100 using ResNet-18 and WRN28-10, respectively. In addition, we implement the softmax+CAM baseline, which applies softmax to the CAMs directly to obtain the mixing mask without using the mask prediction network. We also use the same way to compute the mixing mask for the w/ STN only baseline. The results in Table~\ref{tab:ablation1} show that transforming the input images with softmax+CAM (i.e., the w/ STN only baseline) increases the accuracy the most, while adding the mask prediction network further improves the performance. 

\begin{table}[ht]
  \caption{Test-set accuracy (\%) on CIFAR-10 and CIFAR-100 with different mixing configurations.}
  \label{tab:ablation1}
  \centering
   \begin{tabularx}{\linewidth}{lYYYYYY}
    \toprule
     				& Simple & Mixup & softmax+CAM & w/ STN only & w/ MPN only & \TMIXb \\ \midrule
     	CIFAR-10  		& 95.28 & 95.55 & 95.62$_{\pm0.06}$ & 96.03$_{\pm0.02}$ & 96.02$_{\pm0.01}$ & \textbf{96.40}$_{\pm0.02}$ \\
     	CIFAR-100 & 78.86 & 81.73 & 82.66$_{\pm0.08}$ & 83.79$_{\pm0.04}$ & 83.52$_{\pm0.01}$ & \textbf{84.07}$_{\pm0.05}$ \\
    \bottomrule
  \end{tabularx}
\end{table}

\subsection{Qualitative Analysis}
\begin{figure}[ht]
\centering
\includegraphics[width=0.7\linewidth]{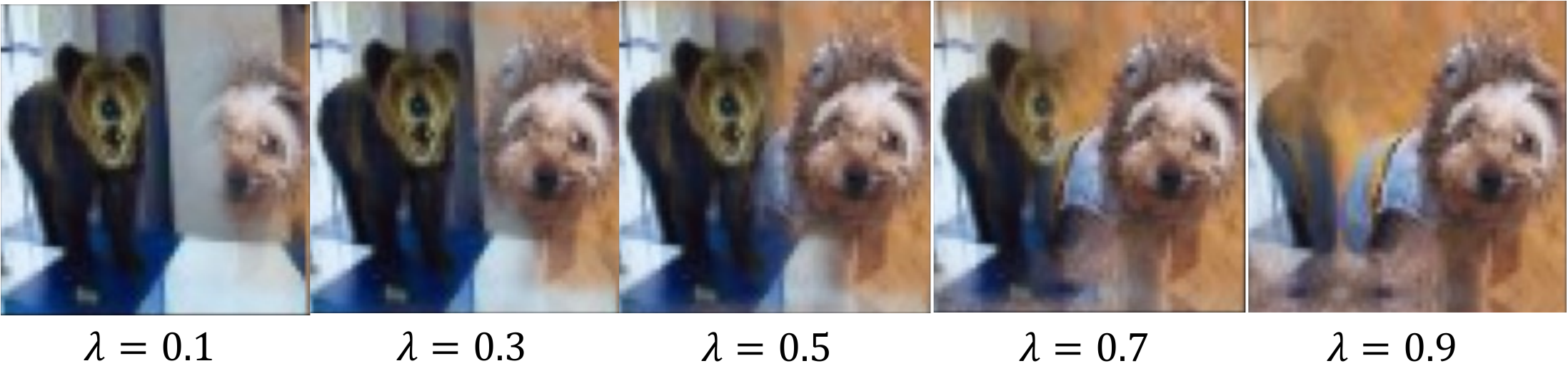}
\caption{Illustration of the mixed outputs with increasing value of the mixing coefficient $\lambda$.}
\label{fig:lambda}
\end{figure}

\noindent We illustrated the learned mixing strategy in Figure~\ref{fig:test1} from Section~\ref{sec:method}. Apparently, the spatial transformer network learns to separate the salient regions of the two input images by squeezing one image to the left and another to the right. Based on the transformed images, the mask prediction network applies the mixing masks that reveal the important areas of the input images. Moreover, the learned temperature value $\tau$ results in a smooth mixing boundary between two objects (see Figure~\ref{fig:test2}). We further validate that \TMIX adapts to the mixing coefficient $\lambda$ well in Figure~\ref{fig:lambda}. The mask prediction model learns to better reveal the second image (the dog) with the increased value of $\lambda$.

\subsection{Mixing More Images}
\label{app:morethan3}
\begin{figure}[ht]
\centering
\includegraphics[width=0.7\linewidth]{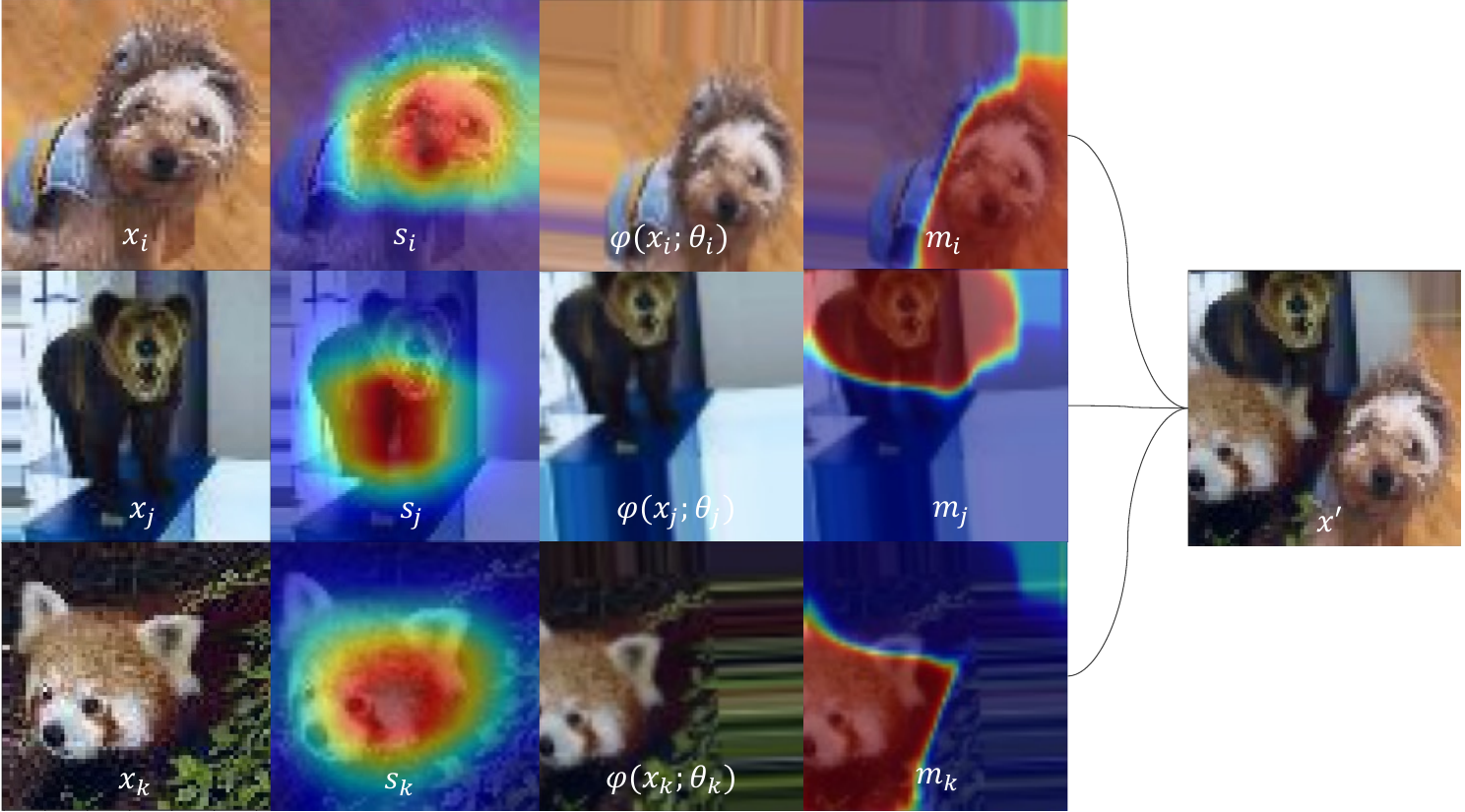}
\caption{Illustration of the mixing strategy for mixing 3 images from Tiny-ImageNet.}
\label{fig:k3_examples}
\end{figure}

\noindent \TMIX can be generalized to mix more than 2 images. Figure~\ref{fig:k3_examples} visualizes the learned mixing strategy when mixing 3 input images on Tiny-ImageNet. The spatial transformer network learns to squeeze the input images to different corners so that the salient regions do not overlap in the mixed results. We compare the end classification performance of \TMIX mixing 3 input images on Tiny-ImageNet with Mixup, CutMix and Co-mixup~\citep{comixup} in Table~\ref{tab:k3_acc}. Our method achieves higher top-1 and top-5 accuracy scores over other baselines. Using the learned mixing strategy with 3 inputs also achieves slightly better performance than 2 inputs on Tiny-ImageNet.

\begin{table}[h]
  \caption{The top-1 / top-5 accuracy (\%) of different baselines and \TMIX with 2 and 3 input images on Tiny-ImageNet.}
  \label{tab:k3_acc}
  \centering
  \begin{tabularx}{\linewidth}{XYYYYYY}
	\toprule
		 Tiny-ImageNet& Simple & Mixup & Cutmix & Co-Mixup & \TMIX  $(k=2)$ & \TMIX  $(k=3)$ \\ \midrule
	top-1	 & 57.23 & 56.59 & 56.67 & 64.15 & 65.72 & \textbf{66.03} \\ 
	top-5	 & 73.65 & 73.02 & 75.52 & -     &  85.0 & \textbf{85.16} \\
    \bottomrule
  \end{tabularx}
\end{table}

\section{Conclusion}
\noindent This paper proposes an automated approach, \TMIXb, to learn transformation and mixing augmentation strategies from data. Based on the class activation maps of the input images, \TMIX employs a spatial transformer network to predict the transformation and a mask prediction network to blend the input images. The mixing module is optimized through self-supervision signals given by a pre-trained teacher network efficiently. Through qualitative and quantitive analysis, we demonstrate the effectiveness of \TMIX in preserving the salient information and improving the end classification performance on multiple datasets under the direct and transfer settings. As our method does not rely on transformations defined in a specific domain, it is beneficial to study whether the method can be modified and deployed to other domains and data modalities in the future.

\bibliography{egbib.bib}

\begin{thebibliography}{35}
\providecommand{\natexlab}[1]{#1}
\providecommand{\url}[1]{\texttt{#1}}
\expandafter\ifx\csname urlstyle\endcsname\relax
  \providecommand{\doi}[1]{doi: #1}\else
  \providecommand{\doi}{doi: \begingroup \urlstyle{rm}\Url}\fi

\bibitem[Bochkovskiy et~al.(2020)Bochkovskiy, Wang, and Liao]{yolov4}
Alexey Bochkovskiy, Chien{-}Yao Wang, and Hong{-}Yuan~Mark Liao.
\newblock {YOLOv4: O}ptimal speed and accuracy of object detection.
\newblock \emph{arXiv preprint arXiv:2004.10934}, 2020.

\bibitem[Cheung \& Yeung(2021)Cheung and Yeung]{modals}
Tsz~Him Cheung and Dit~Yan Yeung.
\newblock {MODALS}: Modality-agnostic automated data augmentation in the latent
  space.
\newblock In \emph{9th International Conference on Learning Representations,
  {ICLR 2021}}, 2021.

\bibitem[Cheung \& Yeung(2022)Cheung and Yeung]{adaaug}
Tsz~Him Cheung and Dit~Yan Yeung.
\newblock {AdaAug}: {L}earning class- and instance-adaptive data augmentation
  policies.
\newblock In \emph{10th International Conference on Learning Representations,
  {ICLR 2022}}, 2022.

\bibitem[Cheung \& Yeung(2023)Cheung and Yeung]{autoda}
Tsz-Him Cheung and Dit-Yan Yeung.
\newblock A survey of automated data augmentation for image classification:
  Learning to compose, mix, and generate.
\newblock \emph{IEEE transactions on neural networks and learning systems}, PP,
  06 2023.
\newblock \doi{10.1109/TNNLS.2023.3282258}.

\bibitem[Chrabaszcz et~al.(2017)Chrabaszcz, Loshchilov, and
  Hutter]{tinyimagenet}
Patryk Chrabaszcz, Ilya Loshchilov, and Frank Hutter.
\newblock A downsampled variant of imagenet as an alternative to the {CIFAR}
  datasets.
\newblock \emph{arXiv preprint arXiv:1707.08819}, 2017.

\bibitem[Cubuk et~al.(2019)Cubuk, Zoph, Man{\'{e}}, Vasudevan, and Le]{aa}
Ekin~D. Cubuk, Barret Zoph, Dandelion Man{\'{e}}, Vijay Vasudevan, and Quoc~V.
  Le.
\newblock {AutoAugment}: Learning augmentation strategies from data.
\newblock In \emph{{IEEE} Conference on Computer Vision and Pattern
  Recognition, {CVPR} 2019}, pp.\  113--123. {IEEE}, 2019.

\bibitem[Cubuk et~al.(2020)Cubuk, Zoph, Shlens, and Le]{randaug}
Ekin~D. Cubuk, Barret Zoph, Jonathon Shlens, and Quoc~V. Le.
\newblock {RandAugment:} {P}ractical automated data augmentation with a reduced
  search space.
\newblock In \emph{IEEE Conference on Computer Vision and Pattern Recognition,
  {CVPR} Workshops 2020}, pp.\  3008--3017. {IEEE}, 2020.

\bibitem[Dabouei et~al.(2021)Dabouei, Soleymani, Taherkhani, and
  Nasrabadi]{supermix}
Ali Dabouei, Sobhan Soleymani, Fariborz Taherkhani, and Nasser~M. Nasrabadi.
\newblock {S}uper{M}ix: {S}upervising the mixing data augmentation.
\newblock In \emph{{IEEE} Conference on Computer Vision and Pattern
  Recognition, {CVPR} 2021}, pp.\  13794--13803. {IEEE}, 2021.

\bibitem[Deng et~al.(2009)Deng, Dong, Socher, Li, Li, and Li]{imagenet}
Jia Deng, Wei Dong, Richard Socher, Li{-}Jia Li, Kai Li, and Fei{-}Fei Li.
\newblock {ImageNet}: {A} large-scale hierarchical image database.
\newblock In \emph{2009 {IEEE} Computer Society Conference on Computer Vision
  and Pattern Recognition {(CVPR} 2009)}, pp.\  248--255. {IEEE}, 2009.

\bibitem[Em et~al.(2017)Em, Gao, Lou, Wang, Huang, and Duan]{pet}
Yan Em, Feng Gao, Yihang Lou, Shiqi Wang, Tiejun Huang, and Ling{-}Yu Duan.
\newblock Incorporating intra-class variance to fine-grained visual
  recognition.
\newblock In \emph{2017 {IEEE} International Conference on Multimedia and Expo,
  {ICME} 2017}, pp.\  1452--1457. {IEEE}, 2017.

\bibitem[Everingham et~al.(2010)Everingham, Van~Gool, Williams, Winn, and
  Zisserman]{voc}
M.~Everingham, L.~Van~Gool, C.~K.~I. Williams, J.~Winn, and A.~Zisserman.
\newblock The pascal visual object classes (voc) challenge.
\newblock \emph{International Journal of Computer Vision}, 88\penalty0
  (2):\penalty0 303--338, June 2010.

\bibitem[Guo et~al.(2019)Guo, Mao, and Zhang]{adamixup}
Hongyu Guo, Yongyi Mao, and Richong Zhang.
\newblock Mixup as locally linear out-of-manifold regularization.
\newblock In \emph{The Thirty-Third {AAAI} Conference on Artificial
  Intelligence, {AAAI} 2019, The Thirty-First Innovative Applications of
  Artificial Intelligence Conference, {IAAI} 2019, The Ninth {AAAI} Symposium
  on Educational Advances in Artificial Intelligence, {EAAI} 2019}, pp.\
  3714--3722. {AAAI} Press, 2019.

\bibitem[He et~al.(2016{\natexlab{a}})He, Zhang, Ren, and Sun]{presnet}
Kaiming He, Xiangyu Zhang, Shaoqing Ren, and Jian Sun.
\newblock Identity mappings in deep residual networks.
\newblock In \emph{Computer Vision - {ECCV} 2016 - 14th European Conference},
  volume 9908, pp.\  630--645. Springer, 2016{\natexlab{a}}.

\bibitem[He et~al.(2016{\natexlab{b}})He, Zhang, Ren, and Sun]{resnet}
Kaiming He, Xiangyu Zhang, Shaoqing Ren, and Jian Sun.
\newblock Deep residual learning for image recognition.
\newblock In \emph{{IEEE} Conference on Computer Vision and Pattern
  Recognition, {CVPR} 2016}, pp.\  770--778. {IEEE}, 2016{\natexlab{b}}.

\bibitem[Hendrycks et~al.(2020)Hendrycks, Mu, Cubuk, Zoph, Gilmer, and
  Lakshminarayanan]{augmix}
Dan Hendrycks, Norman Mu, Ekin~Dogus Cubuk, Barret Zoph, Justin Gilmer, and
  Balaji Lakshminarayanan.
\newblock {AugMix:} {A} simple data processing method to improve robustness and
  uncertainty.
\newblock In \emph{8th International Conference on Learning Representations,
  {ICLR} 2020}, 2020.

\bibitem[Ho et~al.(2019)Ho, Liang, Chen, Stoica, and Abbeel]{pba}
Daniel Ho, Eric Liang, Xi~Chen, Ion Stoica, and Pieter Abbeel.
\newblock {Population Based Augmentation: } efficient learning of augmentation
  policy schedules.
\newblock In \emph{Proceedings of the 36th International Conference on Machine
  Learning, {ICML} 2019}, volume~97, pp.\  2731--2741. {PMLR}, 2019.

\bibitem[Jaderberg et~al.(2015)Jaderberg, Simonyan, Zisserman, and
  Kavukcuoglu]{stn}
Max Jaderberg, Karen Simonyan, Andrew Zisserman, and Koray Kavukcuoglu.
\newblock Spatial transformer networks.
\newblock In \emph{Advances in Neural Information Processing Systems 28: Annual
  Conference on Neural Information Processing Systems 2015}, pp.\  2017--2025,
  2015.

\bibitem[Kim et~al.(2020)Kim, Choo, and Song]{puzzlemix}
Jang{-}Hyun Kim, Wonho Choo, and Hyun~Oh Song.
\newblock {P}uzzle {M}ix: {E}xploiting saliency and local statistics for
  optimal mixup.
\newblock In \emph{Proceedings of the 37th International Conference on Machine
  Learning, {ICML} 2020}, volume 119 of \emph{Proceedings of Machine Learning
  Research}, pp.\  5275--5285. {PMLR}, 2020.

\bibitem[Kim et~al.(2021)Kim, Choo, Jeong, and Song]{comixup}
Jang{-}Hyun Kim, Wonho Choo, Hosan Jeong, and Hyun~Oh Song.
\newblock {C}o-{M}ixup: {S}aliency guided joint mixup with supermodular
  diversity.
\newblock In \emph{9th International Conference on Learning Representations,
  {ICLR} 2021}, 2021.

\bibitem[Krause et~al.(2013)Krause, Deng, Stark, and Fei-Fei]{car}
J.~Krause, Jun Deng, Michael Stark, and Li~Fei-Fei.
\newblock Collecting a large-scale dataset of fine-grained cars.
\newblock In \emph{Second Workshop on Fine-Grained Visual Categorization},
  2013.

\bibitem[Krizhevsky \& Hinton(2009)Krizhevsky and Hinton]{cifar10}
A.~Krizhevsky and G.~Hinton.
\newblock Learning multiple layers of features from tiny images.
\newblock Technical report, University of Toronto, 2009.

\bibitem[Krizhevsky et~al.(2012)Krizhevsky, Sutskever, and Hinton]{alexnet}
Alex Krizhevsky, Ilya Sutskever, and Geoffrey~E. Hinton.
\newblock {ImageNet} classification with deep convolutional neural networks.
\newblock In \emph{Advances in Neural Information Processing Systems 25: 26th
  Annual Conference on Neural Information Processing Systems 2012.}, pp.\
  1106--1114, 2012.

\bibitem[Lim et~al.(2019)Lim, Kim, Kim, Kim, and Kim]{fastaa}
Sungbin Lim, Ildoo Kim, Taesup Kim, Chiheon Kim, and Sungwoong Kim.
\newblock {Fast AutoAugment}.
\newblock In \emph{Advances in Neural Information Processing Systems 32: Annual
  Conference on Neural Information Processing Systems 2019, NeurIPS 2019}, pp.\
   6662--6672, 2019.

\bibitem[Liu et~al.(2021)Liu, Li, Wu, Chen, Wu, Guo, and Li]{automix}
Zicheng Liu, Siyuan Li, Di~Wu, Zhiyuan Chen, Lirong Wu, Jianzhu Guo, and
  Stan~Z. Li.
\newblock {AutoMix: U}nveiling the power of mixup.
\newblock \emph{arXiv preprint arXiv:2103.13027}, 2021.

\bibitem[Maji et~al.(2013)Maji, Rahtu, Kannala, Blaschko, and
  Vedaldi]{aircraft}
Subhransu Maji, Esa Rahtu, Juho Kannala, Matthew~B. Blaschko, and Andrea
  Vedaldi.
\newblock Fine-grained visual classification of aircraft.
\newblock \emph{arXiv preprint arXiv:1306.5151}, 2013.

\bibitem[Nilsback \& Zisserman(2008)Nilsback and Zisserman]{flower}
Maria{-}Elena Nilsback and Andrew Zisserman.
\newblock Automated flower classification over a large number of classes.
\newblock In \emph{Sixth Indian Conference on Computer Vision, Graphics {\&}
  Image Processing, {ICVGIP} 2008}, pp.\  722--729. {IEEE}, 2008.

\bibitem[Qin et~al.(2020)Qin, Fang, Zhang, Liu, Wang, and Wang]{resizemix}
Jie Qin, Jiemin Fang, Qian Zhang, Wenyu Liu, Xingang Wang, and Xinggang Wang.
\newblock Resizemix: Mixing data with preserved object information and true
  labels.
\newblock \emph{arXiv preprint arXiv:2012.11101}, 2020.

\bibitem[Ren et~al.(2015)Ren, He, Girshick, and Sun]{fasterrcnn}
Shaoqing Ren, Kaiming He, Ross~B. Girshick, and Jian Sun.
\newblock Faster {R-CNN:} {T}owards real-time object detection with region
  proposal networks.
\newblock In \emph{Advances in Neural Information Processing Systems 28: Annual
  Conference on Neural Information Processing Systems 2015}, pp.\  91--99,
  2015.

\bibitem[Shorten \& Khoshgoftaar(2019)Shorten and Khoshgoftaar]{da}
Connor Shorten and Taghi~M. Khoshgoftaar.
\newblock A survey on image data augmentation for deep learning.
\newblock \emph{Journal of Big Data}, 6:\penalty0 60, 2019.

\bibitem[Uddin et~al.(2021)Uddin, Monira, Shin, Chung, and Bae]{smix}
A.~F. M.~Shahab Uddin, Mst.~Sirazam Monira, Wheemyung Shin, TaeChoong Chung,
  and Sung{-}Ho Bae.
\newblock {S}aliency{M}ix: {A} saliency guided data augmentation strategy for
  better regularization.
\newblock In \emph{9th International Conference on Learning Representations,
  {ICLR} 2021}, 2021.

\bibitem[Verma et~al.(2019)Verma, Lamb, Beckham, Najafi, Mitliagkas,
  Lopez{-}Paz, and Bengio]{manifoldmixup}
Vikas Verma, Alex Lamb, Christopher Beckham, Amir Najafi, Ioannis Mitliagkas,
  David Lopez{-}Paz, and Yoshua Bengio.
\newblock Manifold {M}ixup: {B}etter representations by interpolating hidden
  states.
\newblock In \emph{Proceedings of the 36th International Conference on Machine
  Learning, {ICML} 2019}, volume~97 of \emph{Proceedings of Machine Learning
  Research}, pp.\  6438--6447. {PMLR}, 2019.

\bibitem[Yun et~al.(2019)Yun, Han, Chun, Oh, Yoo, and Choe]{cutmix}
Sangdoo Yun, Dongyoon Han, Sanghyuk Chun, Seong~Joon Oh, Youngjoon Yoo, and
  Junsuk Choe.
\newblock Cutmix: Regularization strategy to train strong classifiers with
  localizable features.
\newblock In \emph{2019 {IEEE} International Conference on Computer Vision,
  {ICCV} 2019}, pp.\  6022--6031. {IEEE}, 2019.

\bibitem[Zagoruyko \& Komodakis(2016)Zagoruyko and Komodakis]{wrn}
Sergey Zagoruyko and Nikos Komodakis.
\newblock Wide residual networks.
\newblock In \emph{Proceedings of the British Machine Vision Conference 2016,
  {BMVC} 2016}. {BMVA} Press, 2016.

\bibitem[Zhang et~al.(2018)Zhang, Ciss{\'{e}}, Dauphin, and Lopez{-}Paz]{mixup}
Hongyi Zhang, Moustapha Ciss{\'{e}}, Yann~N. Dauphin, and David Lopez{-}Paz.
\newblock {Mixup:} {B}eyond empirical risk minimization.
\newblock In \emph{6th International Conference on Learning Representations,
  {ICLR} 2018}, 2018.

\bibitem[Zhou et~al.(2016)Zhou, Khosla, Lapedriza, Oliva, and Torralba]{cam}
Bolei Zhou, Aditya Khosla, {\`{A}}gata Lapedriza, Aude Oliva, and Antonio
  Torralba.
\newblock Learning deep features for discriminative localization.
\newblock In \emph{2016 {IEEE} Conference on Computer Vision and Pattern
  Recognition, {CVPR} 2016}, pp.\  2921--2929. {IEEE}, 2016.

\end{thebibliography}
\bibliographystyle{tmlr}

\appendix
\section{Appendix}
\subsection{Experiment on CIFAR-10}
We test the effectiveness of \TMIX on the smaller CIFAR-10 dataset and compare the results with Mixup and CutMix. Table~\ref{tab:cifar10} shows that \TMIX achieves higher top-1 accuracy than the other baselines.

\label{app:more_datasets}
\begin{table}[h]
  \caption{Test-set Top-1 accuracy (\%) on CIFAR-10 with ResNet-18 and WRN28-10}
  \label{tab:cifar10}
  \centering
  \begin{tabularx}{\linewidth}{lYYYY}
    \toprule
     & Simple & Mixup & Cutmix & Transform-Mix \\
    \midrule
	 ResNet-18  	 & 95.28 & 95.55 & 96.22 & $\textbf{96.40} {\pm0.026}$ \\
     WideResNet28-10 & 96.13 & 96.9 & 97.13 & $\textbf{97.21} {\pm0.016}$ \\
    \bottomrule
  \end{tabularx}
\end{table}

\subsection{Additional experiment on Tiny-ImageNet}
\label{app:automix}
We compare the classification accuracy with strong baselines using the training protocol prescribed in~\cite{automix}. Table~\ref{tab:automix} shows that our proposed \TMIX is a strong method when compared with the AutoMix baseline on Tiny-ImageNet.

\begin{table}[h]
  \caption{Test-set Top-1 accuracy (\%) for training ResNet-18 network on Tiny-ImageNet.}
  \label{tab:automix}
  \centering
  \begin{tabularx}{\linewidth}{lYYYYY}
  \toprule
	&               Simple & MixUp & CutMix & ManifoldMix & SaliencyMix \\ \midrule
	Tiny-ImageNet & 61.68 & 63.86 & 65.53 & 64.15 & 64.60         \\ \midrule \midrule
	&				PuzzleMix & Co-Mixup  & ResizeMix & AutoMix & \TMIX \\ \midrule
	(Cont.) &			65.81 & 65.92 & 63.74 & 67.33 & \textbf{67.65} \\

\bottomrule
\end{tabularx}
\end{table}
 
\subsection{Comparing Direct and Transfer classification}
\label{app:directvstransfer}
In this section, we compare the effectiveness between direct TransformMix and transfer TransformMix. Specifically, we utilize the mixing module trained on CIFAR-10 to create new augmented images from unseen CIFAR-100 and TinyImageNet. These new images are used to train a WRN-28$\times$10 model on CIFAR-100 and a PreActResNet-18 model on Tiny-ImageNet from scratch using the same training protocol in the direct classification experiments. Since Tiny-ImageNet has a larger image size than CIFAR-10, we resize the CAMs to match the CIFAR-10 images before inputting them into the spatial transformer network. Table~\ref{tab:transfer1} shows that the task models trained with \TMIX images outperform those with Input Mixup, Manifold Mixup, and CutMix by a large margin. The experiment results of the transfer mixing module degrade only slightly when compared to the direct mixing module, which is trained on the target datasets.

\begin{table}[h]
  \caption{Test-set Top-1 accuracy (\%) for training the task network on CIFAR-100 and Tiny-ImageNet datasets using the mixing module learned from CIFAR-10.}
  \label{tab:transfer1}
  \centering
  \begin{tabularx}{\linewidth}{lYYYYYY}
    \toprule
      & Simple & Mixup & Manifold Mixup & CutMix & \TMIX (transfer) & \TMIX (direct) \\
      \midrule
      CIFAR-100 & 78.86 & 81.73 &  82.60 & 82.50 & 84.02$_{ \pm 0.045}$ & \textbf{84.07}$_{\pm 0.05}$ \\
      Tiny-ImageNet & 57.23 & 56.59 & 58.01 & 56.67 & 65.62$_{ \pm 0.29}$ & \textbf{65.72}$_{ \pm 0.16}$ \\
    \bottomrule
  \end{tabularx}
\end{table}

\subsection{Knowledge Distillation}
\noindent Table~\ref{tab:kd} compares \TMIX with SuperMix on the knowledge distillation task. We follow the SuperMix experiment, which uses a WRN-40-2 teacher model to provide the supervision signals for training a ShuffleNetV1 student model on the CIFAR-100 dataset with TransformMix augmentation. Despite the efforts to reproduce the results, we could not fully replicate the baseline performance reported by~\citep{supermix}. In our experiments, \TMIX improves the accuracy by 7.52\% and 0.52\% over the student model trained with no sample-mixing and MixUp, respectively. The gains are comparable to the reported results in SuperMix. This demonstrates the effectiveness of \TMIX beyond standard classification and object detection tasks.

\begin{table}[h]
  \caption{Top-1 test accuracy of the student model trained using knowledge distillation method on CIFAR-100. $^{*}$Reported results from~\citep{supermix}. $^{\dagger}$Reproduced results. }
  \label{tab:kd}
  \centering
  \begin{tabularx}{\linewidth}{lYY}
    \toprule
     & SuperMix$^*$ & \TMIXb$^{\dagger}$ \\
    \midrule
    Teacher Acc. 	& 75.61	& 75.59 \\
    Student Acc.	 	& 70.50	& 69.24 \\
    w/ MixUp 		& 77.44	& 76.24 \\
    w/ Method		& 78.07	& 76.76 \\
    \midrule
    Gain over Student & +7.57 & +7.52 \\                       
    Gain over MixUp & +0.63 & +0.52 \\
    \bottomrule
  \end{tabularx}
\end{table}

\subsection{Sensitivity analysis on network configurations}
\begin{figure}[ht]
    \begin{center}
    \begin{subfigure}[b]{0.22 \linewidth}
        \begin{center}
        \includegraphics[width=\linewidth]{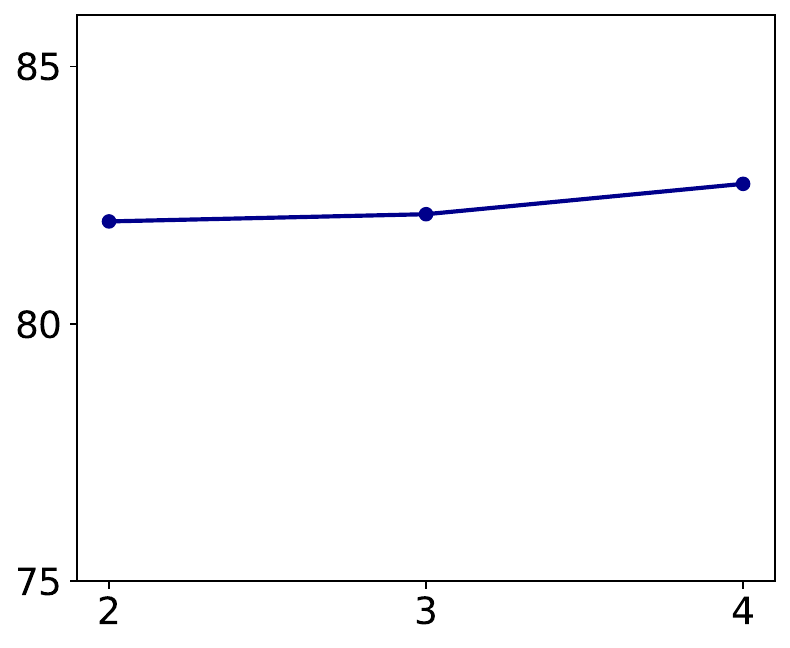}
        \caption{}
        \label{fig:nlayer} 
        \end{center}
    \end{subfigure}
    \,
    \begin{subfigure}[b]{0.22 \linewidth}
        \begin{center}
        \includegraphics[width=\linewidth]{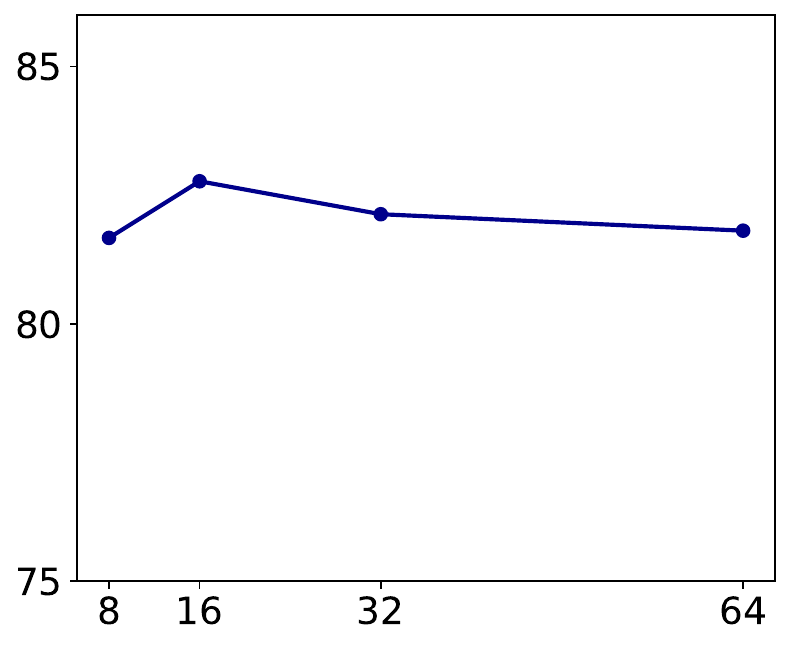}
        \caption{}
        \label{fig:nchannel} 
        \end{center}
    \end{subfigure}
    \,
    \begin{subfigure}[b]{0.22 \linewidth}
        \begin{center}
        \includegraphics[width=\linewidth]{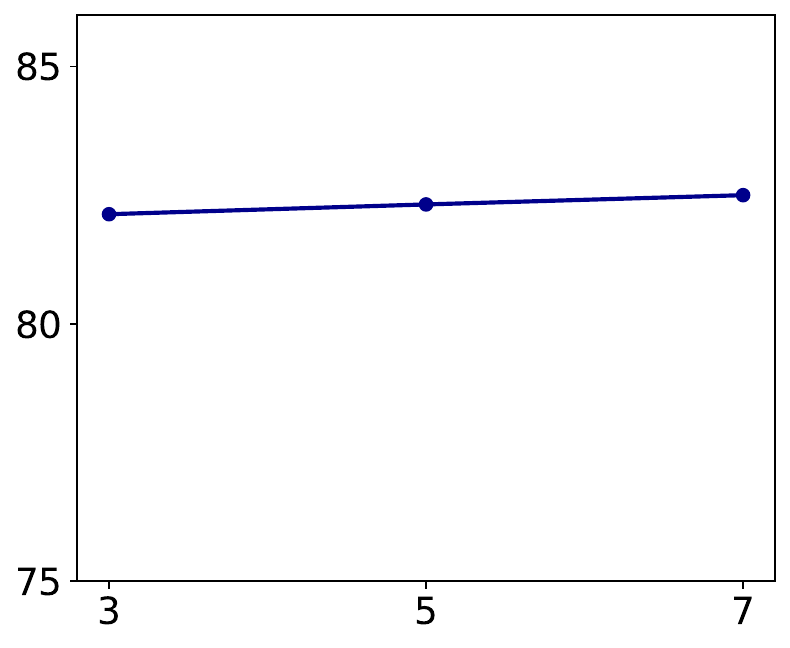}
        \caption{}
        \label{fig:nkernel} 
        \end{center}
    \end{subfigure}
    \,
    \begin{subfigure}[b]{0.22\linewidth}
        \begin{center}
        \includegraphics[width=\linewidth]{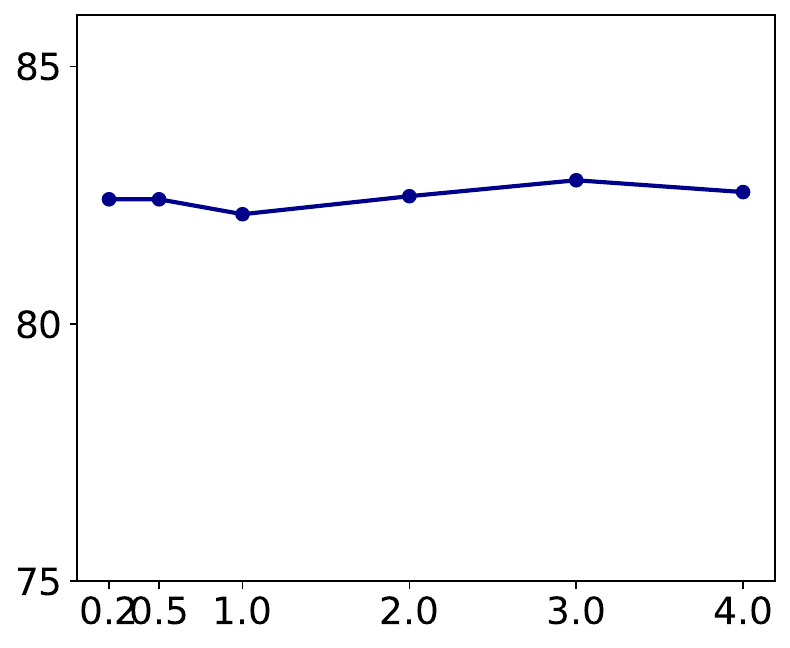}
        \caption{}
        \label{fig:nalpha}
        \end{center}
    \end{subfigure}
    \end{center}
    \caption{Sensitivity to different choices of (a)~number of layers, (b)~number of channels, (c)~kernel size of the mask prediction network and (d)~mixing parameter $\alpha$. The vertical axis shows the top-1 test-set accuracy (\%) and the horizontal axis shows the values of the tested configuration.}
    \label{fig:ila_layers_compare}    
\end{figure}

In \TMIXb, the mask prediction network is implemented as a 3-layer convolutional neural network, where each layer has a channel size of 32 and kernel size of 3x3. We test the sensitivity of our method to the network configurations. Specifically, we compare the end performance of the ResNet-18 models trained on the images created by \TMIX with different implementations of the mask prediction network. The task model is trained on the CIFAR-100 dataset for 200 epochs, and the mask prediction network uses a different number of layers, channel size, and kernel size from \{2, 3, 4\}, \{8, 16, 32, 64\} and \{3, 5, 7\}, respectively. We also explore the effect of using different $\alpha$ values in \{0.2, 0.5, 1, 2, 3, 4\} when sampling the mixing coefficients. The experimental results illustrated in Figure~\ref{fig:nlayer}, \ref{fig:nchannel}, \ref{fig:nkernel} and \ref{fig:nalpha} show that \TMIX is not sensitive to the neural architecture design and the $\alpha$ parameter. It is also worth noting that our method does not require fine-tuning additional parameters on new datasets. This is an advantage over existing methods that require manual specification of additional hyperparameters, such as the label smoothness, data local smoothness, transport cost, and neighbor size in PuzzleMix~\cite{puzzlemix}; the roughness and sparsity coefficients in SuperMix~\cite{supermix}.

\subsection{More visual illustrations of mixed output generated by TransformMix} \label{app:more_visuals}
\begin{figure*}[ht]
\centering
\includegraphics[width=\textwidth]{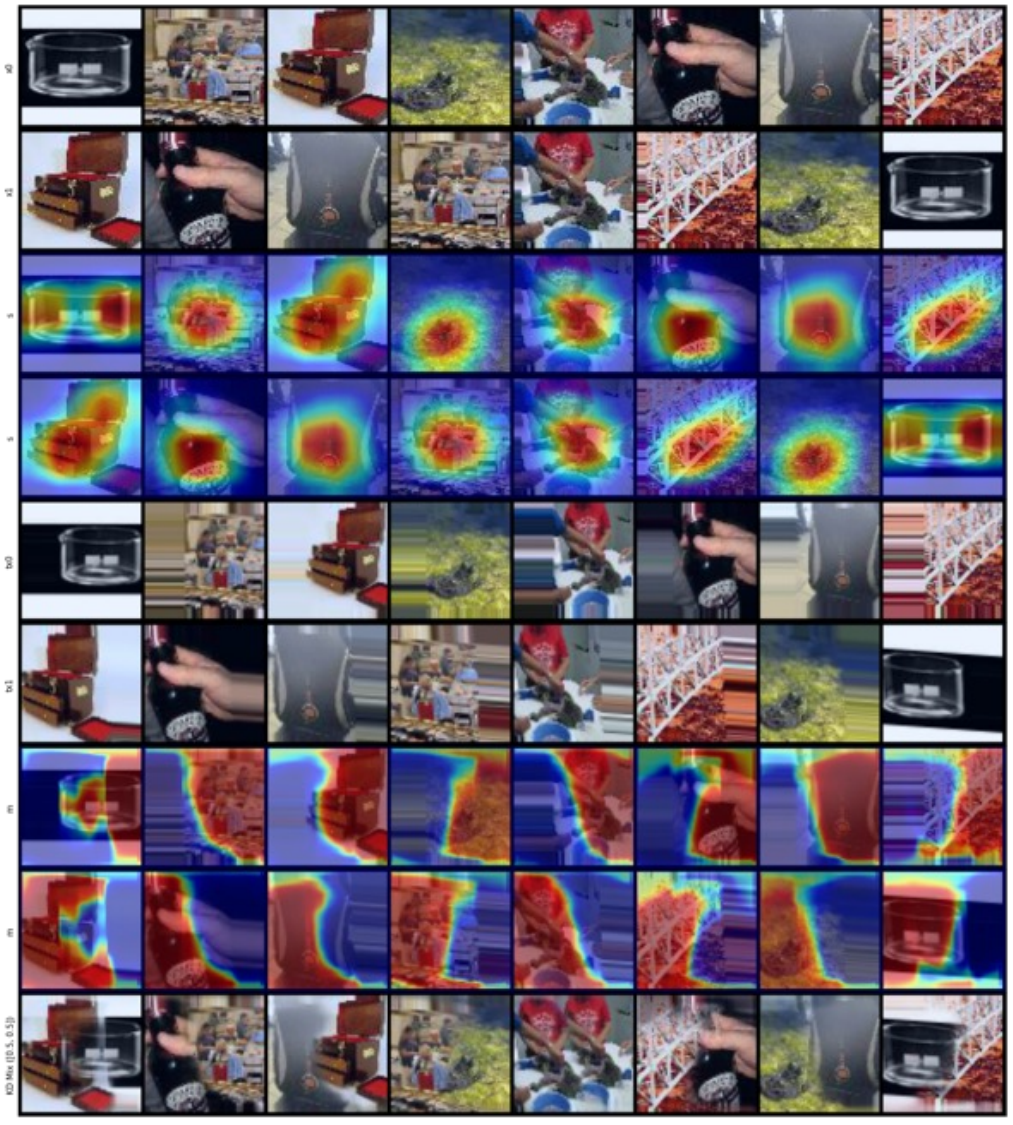}
\caption{More illustration of the mixed outputs from a non-curated set of eight images. From top to bottom, the row indicates input image 1, input image 2, CAM of input image 1, input image 2, transformed input image 1, transformed input image 2, predicted mask of input image 1, the predicted mask of input image 2, and mixed result.}
\label{fig:app1}
\end{figure*}

\begin{figure*}[ht]
\centering
\includegraphics[width=\textwidth]{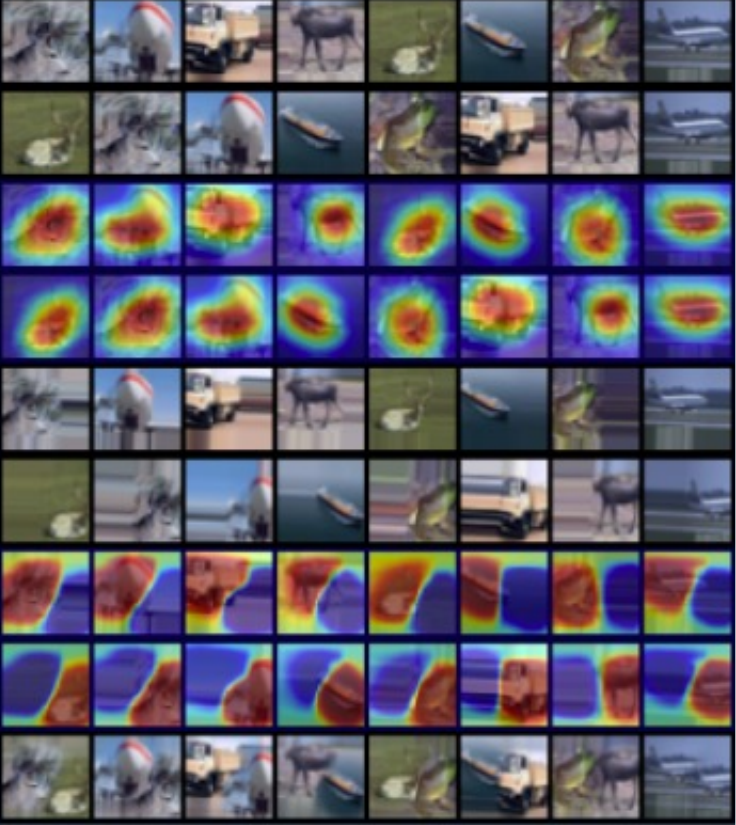}
\caption{More illustration of the mixed outputs from a non-curated set of eight images. From top to bottom, the row indicates input image 1, input image 2, CAM of input image 1, input image 2, transformed input image 1, transformed input image 2, predicted mask of input image 1, the predicted mask of input image 2, and mixed result.}
\label{fig:app2}
\end{figure*}

We present more visualizations of the intermediate results and mixed images generated by \TMIX on different datasets in Figure~\ref{fig:app1} and Figure~\ref{fig:app2}. The illustrations show that \TMIX learns to separate the salient regions of the input images and apply appropriate mixing masks to expose the salient regions on the mixed results.

\end{document}